\documentclass[runningheads]{llncs}

\usepackage{eccv}

\usepackage{eccvabbrv}

\usepackage{graphicx}
\usepackage{booktabs}
\usepackage{makecell}

\usepackage[accsupp]{axessibility}  

\newcommand{\dataset}{FACEMORPHIC}
\newcommand{\fulldataset}{FACEMORPHIC - Neuromorphic Face Dataset}

\usepackage{amssymb}
\usepackage{pifont}
\newcommand{\cmark}{\ding{51}}%
\newcommand{\xmark}{\ding{55}}%

\usepackage[pagebackref,breaklinks,colorlinks,citecolor=eccvblue]{hyperref}

\usepackage{orcidlink}

\begin{document}

\title{Neuromorphic Facial Analysis with Cross-Modal Supervision} 

\titlerunning{Neuromorphic Facial Analysis with Cross-Modal Supervision}

\author{Federico Becattini\inst{1}\orcidlink{0000-0003-2537-2700} \and
Luca Cultrera\inst{2}\orcidlink{0009-0003-2483-9927} \and
Lorenzo Berlincioni\inst{2}\orcidlink{0000-0001-6131-1505} \and
Claudio Ferrari\inst{3}\orcidlink{0000-0001-9465-6753} \and
Andrea Leonardo\inst{2}\orcidlink{} \and
Alberto Del Bimbo\inst{2}\orcidlink{0000-0002-1052-8322}}

\authorrunning{F.~Becattini et al.}

\institute{University of Siena,
Italy
\email{name.surname@unisi.it}\\
\and
University of Florence, Italy \email{name.surname@unifi.it} \and
University of Parma, Italy \email{claudio.ferrari2@unipr.it}}

\maketitle

\begin{abstract}
  Traditional approaches for analyzing RGB frames are capable of providing a fine-grained understanding of a face from different angles by inferring emotions, poses, shapes, landmarks. However, when it comes to subtle movements standard RGB cameras might fall behind due to their latency, making it hard to detect micro-movements that carry highly informative cues to infer the true emotions of a subject. To address this issue, the usage of event cameras to analyze faces is gaining increasing interest. Nonetheless, all the expertise matured for RGB processing is not directly transferrable to neuromorphic data due to a strong domain shift and intrinsic differences in how data is represented.
  The lack of labeled data can be considered one of the main causes of this gap, yet gathering data is harder in the event domain since it cannot be crawled from the web and labeling frames should take into account event aggregation rates and the fact that static parts might not be visible in certain frames.
In this paper, we first present FACEMORPHIC, a multimodal temporally synchronized face dataset comprising both RGB videos and event streams. The data is labeled at a video level with facial Action Units and also contains streams collected with a variety of applications in mind, ranging from 3D shape estimation to lip-reading. We then show how temporal synchronization can allow effective neuromorphic face analysis without the need to manually annotate videos: we instead leverage cross-modal supervision bridging the domain gap by representing face shapes in a 3D space.
  \keywords{Neuromorphic face analysis \and Action Unit Detection \and Cross Modal Supervision}
\end{abstract}

\begin{figure}[t]
    \centering
    \includegraphics[width=\linewidth]{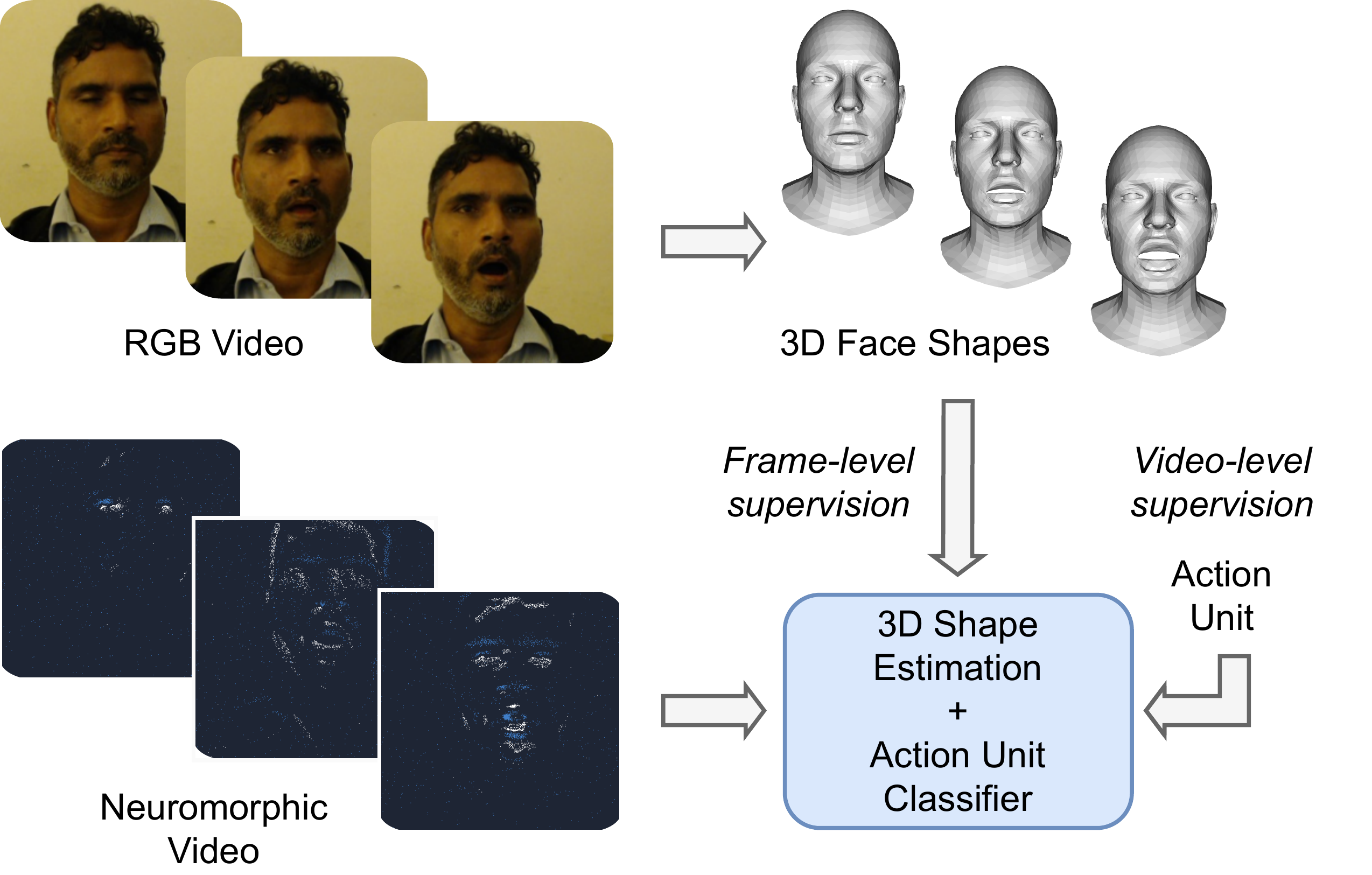}
    \caption{We leverage cross-modal supervision obtainable from temporally synchronized RGB and Event streams to analyze faces using neuromorphic data. By extracting 3D face shape coefficients with standard RGB vision models, we can improve the training of event-based models without additional manual labeling.}
    \label{fig:eyecatcher}
\end{figure}

















\section{Introduction}
\label{sec:intro}

Interpreting human faces is fundamental to many applications, ranging from simple detection up to more complex tasks such as emotion recognition or 3D modeling.
Such importance has resulted in a huge amount of research in this field, which nowadays is supported by a plethora of annotated datasets and open-source models. These provide off-the-shelf tools acting as building blocks for face analysis applications such as face detection \cite{dlib09}, landmark detection \cite{bulat2017far} and gaze estimation \cite{zhang2017mpiigaze}, just to name a few.

Nonetheless, achieving a fine-grained understanding of a face is intrinsically hard: faces continuously produce micro-movements that are the effect of muscle activations and that can happen very suddenly and very quickly. The activation of facial muscles, typically referred to as Action Units, has been largely studied also due to their connection with underlying emotions, to the point that a Facial Action Coding System (FACS) has been created \cite{ekman1978facial}, mapping Action Units to displayed emotions.
The challenge is that micro-expressions and Action Units can have an overall duration as low as 80ms \cite{yan2013fast}. This entails that capturing the evolution of an Action Unit with standard RGB cameras, that operate at 25/30 FPS, might not even be entirely possible.
To effectively model facial dynamics, high-framerate cameras have been used \cite{zhao2023more}, but they require a huge amount of frames to be processed. In this paper, we embrace a different line of research based on neuromorphic vision, which has been gaining increasing interest in the last few years. Neuromorphic sensors, often referred to as event cameras, are biologically inspired sensors that produce asynchronous streams of events rather than synchronous streams of frames. An event is defined as a local change in illumination at a pixel level and can be fired asynchronously at a microsecond rate.
Event cameras have initially gained interest in the field of robotics due to their low consumption, extremely low latency, high dynamic range and absence of motion blur. Nonetheless, a few seminal works on face analysis have recently been proposed \cite{becattini2024neuromorphic, becattini2022understanding, berlincioni2023neuromorphic, berlincioni2024neuromorphic, ryan2023real, shariff2023neuromorphic, bissarinova2023faces}, often focusing on Industry 4.0 applications such as drowsiness estimation \cite{shariff2023neuromorphic} or facial reaction analysis \cite{becattini2022understanding}.

Working with neuromorphic cameras for facial analysis, however, has its shortcomings. Decades of research on RGB images and videos cannot be fully leveraged in the event domain.
In fact, models trained on RGB data do not work on neuromorphic streams due to the heterogeneity of the data \cite{becattini2024neuromorphic}. Even when aggregated as frames \cite{mueggler2017fast, innocenti2021temporal, nguyen2019real}, events exhibit a heavy domain shift that hinders the effectiveness of such models to the point that new architectures must be trained even to address tasks that can be considered close to being solved in the RGB domain (e.g. face detection).
Consequently, the need for annotated neuromorphic datasets is of primary importance, yet just a few, limited ones, exist in the literature and are publicly available~\cite{bissarinova2023faces, berlincioni2023neuromorphic}.
A viable research direction is to convert RGB videos into event streams using simulators \cite{rebecq2018esim, hu2021v2e}. In this way, existing RGB datasets can be leveraged to train models on synthetic data without the need to manually label new footage. Despite a few works recently adopted this approach for detecting faces \cite{becattini2022understanding, berlincioni2023neuromorphic}, converting videos in synthetic data is extremely slow and the generated streams might exhibit spurious events due to compression artifacts \cite{berlincioni2024neuromorphic}. At the same time, fast movements such as micro-expressions captured at slow framerates in video datasets will not appear correctly in the event stream since the signal itself is missing in the source data.


The issue of obtaining high-quality labeled data remains.
To this end, we propose \dataset{}, a new multimodal dataset of subjects performing Action Units recorded with temporally synchronized RGB and Event cameras. Thanks to the temporal alignment, we demonstrate the possibility of deriving a supervision signal directly from the RGB stream by lifting the representation of the face in a camera-invariant reference frame, that is by characterizing it with the coefficients of a 3D Morphable Model, fitted in the RGB domain.

We therefore leverage two distinct sources of supervision: video-level supervision, obtained through manual labeling while collecting the dataset (each video contains a different action unit); and a frame-level cross-domain supervision, derived by applying traditional computer vision models on the RGB frames.
In this sense, the proposed cross-domain supervision resembles the idea of distillation \cite{hinton2015distilling}, where typically a large, well-trained network (\textit{teacher}) is used to supervise the training of a smaller network (\textit{student}).
We show that cross-domain supervision used for regressing 3D face deformation coefficients frame per frame, aids the learning process of an Action Unit classifier. The idea is shown in Fig.~\ref{fig:eyecatcher}.


In summary, the main contributions of our paper are the following:
\begin{itemize}

    \item We propose a learning framework for event data, involving traditional video-level supervision and cross-domain supervision, applied frame-by-frame thanks to information derived from RGB vision modules providing 3D information.

    \item We introduce \fulldataset{}, the first multimodal dataset for action unit detection with temporally synchronized neuromorphic and RGB videos. We collected more than 4 hours of recordings for each modality, including Action Units performed by 64 users.
	
	\item We demonstrate that traditional RGB vision models can be used to supervise event-based tasks when modalities are temporally synchronized. 
\end{itemize}

\section{Related Works}

\textbf{Neuromorphic Vision} Neuromorphic vision describes a set of sensors and acquisition methods based on event cameras which were developed following a novel bio-inspired vision paradigm \cite{delbruckl2016neuro, posch2014retino}. Compared to traditional vision systems, it produces an asynchronous stream of events instead of a synchronous fixed-rate frame sequence. In this domain, an \textit{event} is fired whenever a local change in brightness is detected at an extremely high temporal resolution (in the order of microseconds) with very low latency \cite{lichtsteiner2008asynch}. One of the important properties of these neuromorphic sensors is that they do not output any data unless there is a localized change in brightness, thereby conserving resources and reducing bandwidth consumption \cite{finateu2020back, gallego2020event}. 

These sensors are getting increasingly utilized as they are getting more affordable, and in some domains, such as robotics \cite{mueggler2017event,9813406}, tracking \cite{seok2020robbust,renner2020event}, lip-reading \cite{bulzomi2023end, savran2018energy}, and object detection \cite{DBLP:journals/tim/LiuXYYY23,8593805, magrini2024drone}, the benefits of event cameras can be fully appreciated \cite{gallego2020event,delbruckl2016neuro,ramesh2019dart}.
Besides the aforementioned fields, the unique properties of event cameras find a natural application in the analysis of facial expressions at a high temporal resolution.
Nonetheless, in \cite{berlincioni2024neuromorphic} the human emotion range is analyzed in terms of Valence and Arousal by leveraging an event-data simulator to convert RGB videos. Synthetic approaches such as this do not fully leverage the capabilities of event cameras. 
Despite a rising interest tough, only few datasets focusing on facial dynamics captured using a real event camera are existent in the literature \cite{bissarinova2023faces,savran2020face, moreira2022neuromorphic,  lenz2020event, becattini2022understanding, berlincioni2023neuromorphic, chen2020eddd}.
The authors of~\cite{savran2020face} focus on the task of face pose alignment providing a dataset of 108 videos of several head rotations with varying intensities for a total of 10 minutes of footage.
In \cite{lenz2020event} instead, the presented event-data is collected for eye blink detection. It consists of 48 videos for a cumulative duration of about 13 minutes.
In \cite{becattini2022understanding} the authors present a dataset of 455 videos of facial reactions where the recorded users react to garment images. Finally in \cite{berlincioni2023neuromorphic} the authors collect a dataset composed of paired \textit{RGB - event data} for emotion recognition providing also facial bounding box and landmark annotations in addition to emotion labels.

In this paper we propose a new dataset, \dataset{}, to effectively address face analysis by action unit classification. Comprising more than 4 hours of videos, it is the largest existing dataset related to human facial expressions and emotions, as the datasets collected in  \cite{berlincioni2023neuromorphic} and \cite{becattini2022understanding} have a total extent of 13 and 75 minutes. At the same time, we provide labels covering a set of 24 action units, instead of categorizing videos in binary reaction \cite{becattini2022understanding} or 7 basic emotions \cite{berlincioni2023neuromorphic}.



\textbf{Facial Action Units} The Facial Action Coding System \cite{ekman1978facial} refers to a set of popular facial behavior signs judgment methods. It is an exhaustive anatomical-based system that encodes various facial movements by the combination of basic Action Units (AU). This set of AUs constitutes a sort of alphabet for the, more complex, human face expressions.
Action Units define certain facial configurations caused by the contraction of one or more facial muscles, and they are independent of the interpretation of emotions.
In the human-computer interaction field the use of this taxonomy enabled a large range of applications such as in security settings \cite{Salah2010CommunicationAA,1678017}, clinical studies for pain detection and patient monitoring \cite{LUCEY2012197,Littlewort2007FacesOP}, and in commercial contexts to estimate consumer reaction to products \cite{zhi2020comprehensive,7374704, becattini2021plm}.

\textbf{3D Morphable Models} Since the seminal work of Blanz and Vetter~\cite{blanz2023morphable}, the 3D Morphable Model (3DMM) has been extensively used in the field of face analysis to address a variety of tasks. The 3DMM is a statistical model of shapes, and is usually learned from a set of 3D faces in dense correspondence. Its expressive power depends mostly on the training data, in which direction efforts have been made to build large-scale datasets~\cite{booth20163d} or augmenting the spectrum of possible deformations~\cite{principi2023florence}. Several variants have been proposed for learning a 3DMM, ranging from the standard PCA model~\cite{paysan20093d}, to other solutions based for example on Gaussian process~\cite{luthi2017gaussian}, Dictionary Learning~\cite{ferrari2017dictionary}, multilinear wavelets~\cite{brunton2014multilinear} and so on. Lately, deep networks have been employed as well to learn 3DMMs thanks to their highly non-linear and powerful generation quality~\cite{bouritsas2019neural,ranjan2018generating,tran2018nonlinear,zheng2022imface}. It has been widely used for face reconstruction~\cite{galteri2019deep,gecer2019ganfit,tuan2017regressing,wu2019mvf} or recognition~\cite{hu2016face,an2018deep,koppen2018gaussian,ferrari2015dictionary}, expression and Action Units recognition or generation~\cite{shi2020pose,chang2018expnet,otberdout2023generating,ariano2021action,wang2021improved}. Overall, the related literature reveals the 3DMM still as a state-of-the-art technique for face modeling with several possible applications.

\section{Motivation}
Event cameras can capture illumination changes, and thus motion, at an extremely fast pace. Being able to effectively analyze faces at such a data rate would allow us to precisely characterize expressions and their underlying emotions.
However, face analysis in the neuromorphic domain is proceeding slowly despite increasing interest \cite{becattini2024neuromorphic} motivated by a few preliminary results indicating its usefulness in industrial scenarios \cite{ryan2023real, shariff2023neuromorphic} and its effectiveness over RGB \cite{becattini2022understanding, berlincioni2023neuromorphic}.
Driven by the desire to realize event-based face analysis models, we collected \dataset{} to foster research on this topic. The dataset we collected is temporally synchronized across modalities, meaning that it is possible to obtain different representations of the same face, captured from two sensors - the RGB and the neuromorphic one. As a consequence, any available temporal annotation can be transferred from one media to another.

The same cannot be said for spatial annotations, such as bounding boxes or facial landmarks, since the sensors are uncalibrated and the subject can sit freely in front of the cameras. Working under the assumption of uncalibrated cameras though opens up the problem of precisely annotating data. Spatially labeling faces in RGB frames is almost trivial: off-the-shelf face detectors and landmark detectors can nowadays be effectively used to gather labels automatically, without any manual labor. Conversely, such vision models cannot be directly applied to event frames.
Becattini et al. \cite{becattini2024neuromorphic}, have recently shown that the outputs of RGB models, such as face and landmark detectors, are likely to fail on neuromorphic data, pointing out that when a small amount of motion is present, no meaningful output is obtainable, whereas when faces are sufficiently visible in the event frame, some low-confidence and imprecise detection might be still be gathered. In both cases, such models are unlikely to be of any practical use.


In virtue of these considerations, our goal is to leverage temporally synchronized modalities to transfer automatically generated labels from RGB frames onto events, exploiting a characterization of the face that goes beyond a spatial reference system grounded on either the RGB or the event frame. To this end, we will infer the coefficients of a 3D Morphable Model for event-based frames, bypassing the generation of bounding boxes and facial landmarks, that are usually required to obtain such information.
In this way, by lifting the annotations from the frame to a pose-agnostic 3D representation we can effectively describe the shape of the face and label event streams with a cross-modal supervision.


\section{The \dataset{} Dataset}
In this section, we present \fulldataset{}, that we collected for our experiments. To the best of our knowledge it is the first multimodal RGB and Event dataset for Facial Action Unit classification. All the videos in the dataset are temporally synchronized across modalities and are recorded with a commercial USB RGB camera and a Prophesee Evaluation Kit 4 (EVK4), equipped with the IMX646 neuromorphic sensor. The RGB camera and the event camera have different resolutions, respectively of $640 \times 480$ and $1280 \times 720$.
Each data acquisition session was performed by recording with the two cameras the following for each user: (i) an initial recording, where the user can freely interact with the environment, speak and look around; (ii) separate recordings of the user performing 24 different Action Units - each recording is repeated twice; (iii) four recordings of the subject reading a short sentence drawn at random.
Overall, we collected a total of 3148 videos, corresponding to 4.13 hours of recording for each modality. In the dataset, 64 users are present (16 females and 48 males), ranging from age 18 to 67. 

The 24 Action Units recorded in the dataset include 18 micro-actions related to face muscle activations plus 6 macro-actions involving head movements. All the Action Units have been selected among the Facial Action Coding System (FACS) \cite{ekman1978facial} and were chosen in order to include the Action Units that are usually studied in the vision literature.
In particular, we include all the Action Units used in the popular DISFA\footnote{Action Units 1, 2, 4, 6, 9, 12, 25, 26} \cite{mavadati2013disfa} and BP4D\footnote{Action Units 1, 2, 4, 6, 7, 10, 12, 14, 15, 17, 23, 25} \cite{zhang2014bp4d} datasets. In addition we included also the ones related to eye movements (AU 43 and 45) and head movements (AU 51, 52, 53, 54, 55, 56).
The complete list of Action Units present in \dataset{} is: 1, 2, 4, 6, 7, 9, 10, 12, 14, 15, 17, 23, 24, 25, 26, 27, 43, 45, 51, 52, 53, 54, 55, 56.
In Tab. \ref{tab:datasets} we present a comparison between \dataset{} and existing facial neuromorphic datasets from the literature.
Differently from all the other datasets, \dataset{} is the only dataset providing synchronized RGB+Event data. This enables cross-modal supervision, thus allowing us to learn complex facial dynamics without costly annotation procedures.
It must be noted that most datasets are either extremely small or are recorded with low-resolution sensors. The only existing large-scale dataset is FES \cite{bissarinova2023faces}, which nonetheless addresses only face detection at a low resolution and does not come with RGB data.
\dataset{} is going to be publicly released for research purposes. The release will include also facial bounding boxes and 3D landmarks estimated from the RGB frames using Face Alignment \cite{bulat2017far} and the face shape coefficients for the 3D Morphable Model fitted on the landmarks, as described in Sec. \ref{sec:3dmm}. We are also releasing the sentences read by the users in the data acquisition process. Making subjects read sentences was intended as a way to let users move their faces in a natural way, but since some interest in lip reading with event cameras has been shown in the literature \cite{tan2022multi, bulzomi2023end}, we release the annotations and we leave this to further investigation in future work. In all our experiments we defined an 80-20 split between train and test videos.

\newcommand{\xmarktable}{-}

\begin{table*}[t]
\caption{Comparison with other neuromorphic face datasets. }
    \resizebox{\linewidth}{!}{
		\begin{tabular}{l|ccccccccc}
         Dataset & Year & Videos & Duration & Users  & Resolution & Task & Open Source & ~~~RGB~~~ & Synch. \\
         \hline
         Savran et al. \cite{savran2020face} & 2020 & 108 & 10 min. & 18 & 304$\times$204 & Face Pose Align. & \xmarktable & \xmarktable & \xmarktable  \\
         Lenz et al. \cite{lenz2020event} & 2020 & 48 & 13 min. & 10 & 640$\times$480 & Face Detection & \xmarktable & \xmarktable & \xmarktable  \\
        NEFER \cite{berlincioni2023neuromorphic} & 2023 & 609 & 13 min. & 29 & 1280$\times$720 & Emotion Class. & \ding{51} & \ding{51} & \xmarktable \\
        Savran et al. \cite{savran2018energy} & 2018 & 360 & 28 min. & 18 & 304 $\times$ 240 & Voice Activity Det. & \xmarktable & \xmarktable & \xmarktable \\
         Becattini et al. \cite{becattini2022understanding} & 2022 & 455 & 75 min. & 25 & 640$\times$480 & Reaction Class. & \xmarktable  & \ding{51} & \xmarktable \\

         Chen et al. \cite{chen2020eddd}  & 2020 & 260 & 86 min. & 26 & 346$\times$260 & Driving Monitoring & \ding{51} & \ding{51} & \xmarktable \\
         Moreira et al. \cite{moreira2022neuromorphic} & 2022 & 432 & 180 min. & 40 & - & Identity Recognition & \xmarktable & \xmarktable & \xmarktable \\

         Tan et al. \cite{tan2022multi} & 2022 & 200 & 231 min. & 40 & 346$\times$260 & Lip Reading & \ding{51} & \ding{51} & \xmarktable \\
         FES \cite{bissarinova2023faces}~ & 2023 & 3889 & 689 min. & 73 & 408$\times$360 & Face Detection & \ding{51} & \xmarktable & \xmarktable \\
         \hline
         FACEMORPHIC & 2024 & 3148 & 248 min. & 64 & 1280$\times$720 & Action Unit Class. & \ding{51} & \ding{51} & \ding{51}
    \end{tabular}
    }
    \label{tab:datasets}
\end{table*}

\section{Cross-Modal Labeling with 3D Face Shapes}
In order to effectively train an Action Unit classifier with event data, we rely on video-level supervision (e.g. the AU class label) as well as a cross-modal supervision at frame level. Such supervision comes from face shape coefficients estimated from the RGB with a 3D Morphable Model. The 3D shape can be directly transferred onto event frames since modalities are temporally synchronized and since the face shape is invariant to the point of view (i.e. the camera position). In the following we first provide details about the 3D Morphable Model used to estimate face shapes and we then motivate the annotation transfer across modalities.

\subsection{3D Morphable Model}
\label{sec:3dmm}
The 3D Morphable Model (3DMM) is a statistical deformation model for 3D faces, firstly proposed by Blanz and Vetter~\cite{blanz2023morphable}. A 3DMM is built by learning a low-dimensional space from a set of densely registered 3D faces. The learned basis vectors are used to parameterize the shape (and optionally texture) space and synthesize new faces as:

\begin{equation}
    \mathbf{S} = \mathbf{T} + \mathbf{C}\mathbf{\alpha}
    \label{eq:3dmm}
\end{equation}
where $\mathbf{S}\in \mathbb{R}^{N \times 3}$ is a 3D face of $N$ vertices, $\mathbf{T}\in \mathbb{R}^{N \times 3}$ is a template 3D face, $\mathbf{C}\in \mathbb{R}^{3N \times K}$ are the shape bases and $\alpha \in \mathbb{R}^K$ are the deformation coefficients.

Depending on the variability of the 3D faces in the training dataset, different bases of \textit{deformation components} can be learned. For example, the Basel Face Model~\cite{paysan20093d} is built from a set of 200 faces in neutral expression, and the learned deformations encode structural facial traits \textit{e.g.} thin/large head, feminine/masculine etc. Differently, other methods such as the DL-3DMM described in~\cite{ferrari2015dictionary} or the FLAME~\cite{li2017learning} model are learned from mixed neutral and expressive faces, so the model can also replicate expression related deformations \textit{e.g.} mouth opening, eyebrow raising.

\begin{figure}[!t]
    \centering
    \includegraphics[width=\linewidth]{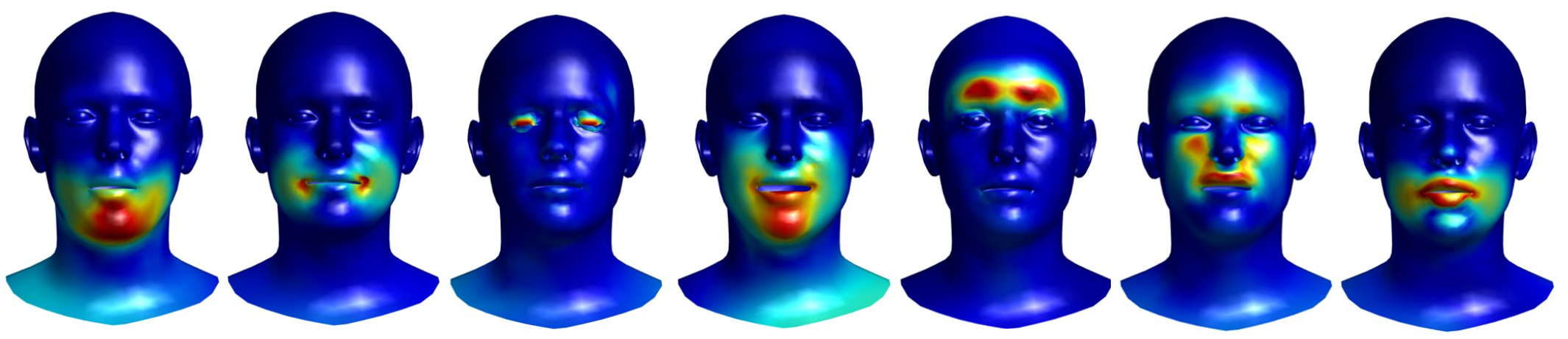}
    \caption{Example of AU-3DMM components learned from the D3DFACS dataset. The heatmaps show the spatial extent of the deformation (\textcolor{red}{red}=high, \textcolor{blue}{blue}=no deformation). The learned components capture AU specific facial movements.}
    \label{fig:AU-3DMM}
\end{figure}

For the purpose of this work, we build two separate 3DMMs, one for encoding structural identity deformations, and the other specific for encoding action-units activations. To this aim, we used the VOCASET~\cite{VOCA2019} and D3DFACS~\cite{Cosker2011AFV} datasets: the former includes 3D sequences of 12 actors performing facial expressions, while the latter 3D sequences of 10 actors performing AU activations. All meshes share the same (FLAME~\cite{li2017learning}) topology. We build the identity model $\mathbf{C}_I \in \mathbb{R}^{3N \times 22}$ (ID-3DMM) from the 22 joined actors, using only samples in neutral expression to learn the PCA space. To build the AU model $\mathbf{C}_{AU} \in \mathbb{R}^{3N \times K}$ (AU-3DMM), we first compute AU-specific deformation offsets between expressive and neutral scans; then, we learn the the deformation components by applying the DL-3DMM algorithm of~\cite{ferrari2015dictionary} on such offsets. Various number of components $K$ have been tested. Some examples of how the learned AU-3DMM components can capture AU activations are shown in Fig.~\ref{fig:AU-3DMM}.

Given the above models, we can then fit the 3DMM to the RGB frames, so to obtain identity and AU-specific deformation coefficients, $\alpha_I$ and $\alpha_{AU}$.  

\subsection{Two-step 3DMM Fitting}\label{subsec:3dmm_fitting}
In order to recover AU-specific deformation coefficients, it is first necessary to remove structural shape information related to the identity. To this aim, we perform a two-step 3DMM fitting, similar to~\cite{ferrari2018rendering}. Given that we are not interested in recovering accurate a 3D face reconstruction but only in capturing facial movements, we apply a landmark-based fitting algorithm. We chose to exploit the method in~\cite{ferrari2015dictionary} as it is extremely fast (solved in closed-form) and sufficiently accurate in modeling facial deformations. 

First, $68$ facial landmarks $\mathbf{l} \in \mathbb{R}^{68 \times 3}$ are detected from the RGB frames using the method in~\cite{bulat2017far}, which also provides an approximate $z$ coordinate for each landmark. A corresponding set of 3D landmarks $\mathbf{L}_T \in \mathbb{R}^{68 \times 3}$ is labeled on a 3D face template $\mathbf{T} \in \mathbb{R}^{5023 \times 3}$ in FLAME topology. Given the detected landmarks and the template landmarks, the fitting is performed by first estimating an orthographic camera model from the landmark correspondence as
   $ \mathbf{A} = \mathbf{l} \cdot \mathbf{L}_T^\dagger$,
%
where $\mathbf{A} \in \mathbb{R}^{2 \times 3}$ is the camera matrix that contains 3D rotation, scale and shear parameters, and $\mathbf{L}_T^\dagger$ indicates the pseudo-inverse matrix. Then, we estimate the 2D translation $\mathbf{t} \in \mathbb{R}^{68 \times 2}$ as $\mathbf{t} = \mathbf{l} - \mathbf{A} \cdot \mathbf{L}_T$. Finally, the deformation coefficients $\alpha$ are estimated by minimizing the projection error between the detected landmarks and the back-projected template landmarks. This problem is cast as a regularized ridge-regression problem:

\begin{equation}
    \min_{\alpha} \left \| \mathbf{l} - \mathbf{C}(\mathbf{A} \cdot \mathbf{L}_T + \mathbf{t} )\alpha \right \|_2^2 + \lambda \left \| \alpha \right \|_2
    \label{eq:fitting}
\end{equation}
which has a closed form solution as shown in~\cite{ferrari2015dictionary}. The parameter $\lambda$ controls the intensity of the deformation and serves to avoid excessive deformations of the template. A new 3D face is then synthesized using Eq.~\eqref{eq:3dmm}. 

\textbf{Identity Model Fitting}\label{subsubsec:id_fit}
Assuming the first frame of each recorded video portrays the subject in neutral expression, we use the ID-3DMM to reconstruct and identity-specific 3D face on this frame. This is simply done by using the components $\mathbf{C}_I$ in Eq.~\eqref{eq:fitting}. As a result, we estimate identity-specific coefficients $\alpha_I$, so the new shape can be obtained as $\mathbf{S}_I = \mathbf{T} + \mathbf{C}_I\alpha_I$.

\textbf{Estimating AU-specific deformation coefficients}
We use the estimated identity shape $\mathbf{S}_I$ to fit the AU model $\mathbf{C}_{AU}$ on all the subsequent frames of the RGB video. This strategy is intended to explicitly disentangle structural and AU-related shape deformations. Ideally, given that $\mathbf{S}_I$ captures the subject identity traits, if we use this 3D shape to fit the frames where the subject performs AU activations, the corresponding deformation coefficients $\alpha_{AU}$ should only capture the Action Units. To this aim, we repeat the process of Sec.~\ref{subsec:3dmm_fitting} yet this time using $\mathbf{S}_I$ in place of the template $\mathbf{T}$, and the AU-3DMM $\mathbf{C}_{AU}$ in place of $\mathbf{C}_I$. Hence, we collect a set of AU-specific coefficients $\alpha_{AU}$ for each frame.


\subsection{Cross-Modal Labeling}
Once face shapes, represented by the coefficients $\alpha_{AU}$, have been obtained, we can map them to event data without further manual annotation.
In this paper, we simplify data transfer across modalities by generating event frames from raw events using an accumulation of 33ms. This yields event videos at 30 FPS, i.e. the same frame-rate of the RGB videos. Associating the coefficients to event frames thus is trivial, as we obtain the same number of frames in both modalities.

Two important matters have to be taken into account. 
First, since we strive to model micro-movements as fast as Action Units, a finer frame-rate could be desirable. In this case, frame association can be done by searching for the frame with the nearest timestamp. The annotation will not be dense, meaning that only 30 frames in each second will be annotated. The remaining frames will either be left without direct supervision or can be labeled by interpolating the temporally adjacent coefficients. We leave this investigation for future research.
Second, we argue that adopting a frame-rate of 30 FPS does not affect the information collected by the event camera. In fact, if RGB cameras create frames by making a snapshot of the current intensity values for each pixel, event frames accumulate all the temporal information within the last $\Delta t=33ms$. This has the advantage of not increasing the frame number compared to RGB (hence, we have no increase in the amount of computation) while still being able to capture motions that happen at timesteps that are not multiples of the frame rate. 
To aggregate events, we use the Periodic Frame Generation Algorithm implemented in the Prophesee SDK\footnote{https://docs.prophesee.ai/stable/concepts.html\#generating-frames-from-cd-events}.


\section{Neuromorphic Action Unit Classification}
\label{sec:problem_statement}
To classify Action Units from event streams, we propose a multi-task model optimizing two losses. First, we minimize a classification loss $\mathcal{L}_{AU}$ for the main task at a video level, i.e. optimizing the probability of each class after having observed the whole sequence. The second loss $\mathcal{L}_{\alpha}$ is optimized for every frame in the video sequence, regressing the coefficients $\alpha_{AU}$ that define the face shape.
More formally, our training setting is the following. Given a set of videos composed of $N$ event frames $F_t=F^1,..., F^N$, we supervise our model with a sequence of $N$ face shape coefficients $\alpha_{AU}^{*t}, t=1,..., N$ and a video class label $c^*_{AU}$, indicating the Action Unit performed in the video.
The resulting loss is therefore $\mathcal{L} = \mathcal{L}_{AU} + \lambda \mathcal{L}_{\alpha}$, where $\lambda$ is a hyperparameter balancing the two losses.
We use for classification a cross-entropy loss $\mathcal{L}_{AU} = -\sum_{i=1}^C c^*_{AU}log(c_i)$, and for regression a Mean Squared Error loss averaged over each frame $\mathcal{L}_{\alpha} = \frac{1}{N}\sum_{t=1}^N||\alpha_{AU}^t - \alpha_{AU}^{*t}||_2^2$, where $c_i$ is the logit for the $i-th$ Action Unit, C is the number of Action Units and $\alpha_{AU}^t$ is the 3D face shape coefficient vector predicted for frame $t$.

In principle, any model capable of processing videos can be used. We tested several alternatives namely (a) ResNet18+LSTM; (b) ResNet18+Transformer; (c) I3D.
The first two architectures leverage a ResNet18 model, pre-trained on ImageNet, that acts as a backbone extracting 1024-dimensional features. We observed that, even if the model was trained on RGB data, the classification network still benefits from the pre-training. We freeze the convolutional part of the model, finetuning the fully connected layers and connecting them to either an LSTM or Transformer layer.
For the LSTM model, we use three layers with hidden size of 256. The final hidden state is then fed for every timestep to a regression head composed of two fully connected layers with size 128 and 32 as the number of components describing the face shape to be regressed $\alpha_{AU}$. Similarly, a classification head with two fully connected layers with sizes 128 and 24, followed by a sofmax activation, generates a probability distribution over Action Units. This head is fed with the final hidden state of the LSTM, after the whole sequence has been processed.
The transformer model operates in a similar way. The ResNet18 outputs for each frame are fed to a transformer encoder, yielding a sequence of outputs, which are then fed to regressors with the same structure as the ones in the LSTM model. Along with the input tokens of the transformer, we fed a CLS token, which, after being processed by the encoder, we use as input for a classification head.

The I3D model instead, follows a different structure. We use a single branch Inception model with Inflated 3D convolutions \cite{carreira2017quo}. Here, the whole sequence of event frames is stacked together as a 3D tensor and processed to obtain a final 256-dimensional feature. As in the previous models we use an Action Unit classification head and a face shape regression head, however, since we do not process frames individually, we directly generate the concatenation of all the 3D coefficients, i.e. the final fully connected layer has an output dimension of $L \times 32$, where $L$ is the sequence length, that we fix to 75 frames (2.5 seconds).
All fully connected layers except the final ones have ReLU activations in all the models. The models are trained with Adam using a learning rate of 0.001.

\newcolumntype{H}{>{\setbox0=\hbox\bgroup}c<{\egroup}@{}}
\begin{table}[t]
\caption{AU classification from 3D face deformation coefficients $\alpha_{AU}$.}
	\centering
	\resizebox{0.85\linewidth}{!}{
		\begin{tabular}{@{}lHccc@{}}
			\toprule
			Model & $\alpha_{AU}$ representation & ~~~~~~Accuracy~~~~~~ & ~~~~~~top 3 Accuracy~~~~~~ & ~~~~~~top 5 Accuracy~~~~~~\\
			\midrule
			Transformer & Sequence & \textbf{69.27} & \textbf{82.76}	& \textbf{87.11}\\
			LSTM & Sequence & 50.34 & 82.36 & 86.66\\
			\bottomrule
		\end{tabular}
	}
	\label{tab:AUalphas}
\end{table}

\section{Experiments}

\textbf{Classification from Face Shapes}
To assess the quality of our proposed dataset, Tab. \ref{tab:AUalphas}, shows the results of a control experiment without employing event data. 
First, landmarks are extracted from the frames through face alignment \cite{bulat2017far}, and subsequently, the model described in Sec. \ref{subsec:3dmm_fitting} computes the deformation coefficients denoted as $\alpha_{AU}$. Finally, we feed the estimated coefficients to an Action Unit classifier.
In Tab. \ref{tab:AUalphas}, models are trained to classify Action Units in videos, utilizing alpha coefficients for each frame.

Treating the coefficients as a sequence, we trained both a Transformer (2 encoder layers; 2 decoder layers; 2 heads) and an LSTM model (1 layer with hidden size 256). Interestingly, the Transformer model achieved the best results ($69.27\%$ accuracy in Tab. \ref{tab:AUalphas}).
This control experiment demonstrates that the collected videos carry a sufficiently informative signal to effectively estimate 3DMM coefficients using models pre-trained on several datasets such as \cite{bulat2017far, Cosker2011AFV, VOCA2019}, yielding excellent performance in Action Unit classification. Given the non-trivial nature of Action Unit classification, training models from scratch poses considerable challenges.

\begin{table}[t]
\caption{Comparison of action unit classification accuracy for models trained with event data or RGB data.}
	\centering
	\resizebox{0.8\textwidth}{!}{
		\begin{tabular}{@{}lccccc@{}}
			\toprule
			Model~~~ & ~~~Mod~~~ & ~~~Accuracy~~~ & ~~~top 3 Accuracy~~~ & ~~~top 5 Accuracy~~~\\
			\midrule
			ResNet18+LSTM & Event& 46.23 & 62.91 & 68.58\\
			ResNet18+Transf. & Event&  31.74 & 40.63 & 52.08\\
			I3D & Event& \textbf{47.08} & 69.58 & 80.66\\
			\midrule
            ResNet18+LSTM  &  RGB& 25.65 & 38.86 & 46.12 \\
             ResNet18+Transf.  &RGB& 4.16 & 12.50 & 20.83\\
            I3D  &RGB& 29.86 & \textbf{72.00} & \textbf{82.05}\\
\bottomrule
		\end{tabular}
	}
	\label{tab:eventVSrgb}
\end{table}

\noindent\textbf{Event vs RGB Comparison}
To motivate the usage of neuromorphic data for action unit classification, we present in Tab. \ref{tab:eventVSrgb} a comparison between the models presented in Sec. \ref{sec:problem_statement}, trained with RGB or event data. It clearly emerges that the models trained with neuromorphic data outperform their RGB counterparts. The ResNet18+Transformer model struggles the most to address the task. We impute this to the fact that the transformer layer is trained from scratch and that it would likely require a larger amount of samples to be trained effectively. Surprisingly, the model when trained on RGB data does not learn effectively, yielding a top-1 accuracy which is equal to a random guess. This does not happen with event data, as the model reaches 31.74\%, which however is still lower than ResNet18+LSTM and I3D. We believe that using event data helps training as events let the model focus on cues that are relevant to the task (the motion of facial parts).
These findings confirm the importance of modeling facial dynamics with neuromorphic cameras rather than RGB data.

\begin{table*}[t]
\caption{Accuracy for different models, average and per Action Unit.}
\centering
\resizebox{1.0\textwidth}{!}{
\begin{tabular}{@{}l|ccc|cccccccccccccccccccccccc@{}}
\toprule
\multicolumn{4}{c}{}  & \multicolumn{20}{c}{Accuracy (\%)}\\ 
\midrule
Model & ~~Acc~~ & ~~Top3~~ & ~~Top5~~ & ~~1~~~ & ~~2~~~ & ~~4~~~ & ~~6~~~ & ~~7~~~ & ~~9~~~ & ~~10~~ & ~~12~~ & ~~14~~ & ~~15~~ & ~~17~~ & ~~23~~ & ~~24~~ & ~~25~~ & ~~26~~ & ~~27~~ & ~~43~~ & ~~45~~ & ~~51~~ & ~~52~~ & ~~53~~ & ~~54~~ & ~~55~~ & ~~56~~ \\ \hline


ResNet18+LSTM & \textbf{50.21} & 70.71 & \textbf{81.17} & \textbf{40.9} & 30.8 & \textbf{53.0} & 34.8 & 45.6 & 18.3 & 29.6 & 6.1 & 42.9 & \textbf{17.6} & 34.5 & 0.0 & \textbf{23.4} & 36.4 & 39.6 & 65.5 & \textbf{57.1} & \textbf{73.6} & \textbf{94.0} & \textbf{85.1} & 92.3 & \textbf{100} & \textbf{87.3} & 98.6 \\

ResNet18+Transf. & 43.35   &   65.59 & 80.05 & 20.3 & \textbf{53.4} & 46.5 & 28.8 & 46.3 & \textbf{43.3} & 10.8 & 1.0 & 26.2 & 13.2 & 28.1 & 30.5 & 4.4 & \textbf{38.0} & 11.0 & 47.5 & 36.9 & 32.3 & 90.8 & 66.5 & \textbf{100} & 84.0 & 80.7 & \textbf{100} \\

I3D & 49.58 & \textbf{71.08} & 79.41 & 29.6 & 45.1 & 40.1 & \textbf{40.0} & \textbf{50.7} & 13.4 & \textbf{62.7} & \textbf{60.2} & \textbf{43.3} & 1.7 & \textbf{41.4} & \textbf{36.0} & 4.2 & 37.2 & \textbf{65.6} & \textbf{70.9} & 39.1 & 62.6 & 57.3 & 84.4 & 63.9 & 90.2 & 75.8 & 77.1 \\
\bottomrule

\end{tabular}
}
\label{tab:AU_acc}
\end{table*}

\noindent\textbf{Results with Cross-Modal Supervision}
Tab. \ref{tab:AU_acc}  presents the Action Unit classification outcomes using event data for the multi-task models presented in Sec. \ref{sec:problem_statement}. For each model, the accuracy for all Action Units is also reported. Notably, ResNet18+LSTM is the model that achieves the best performance, obtaining a Top 1 accuracy of 50.21\%, Top 3 Accuracy of 70.71\%, and Top 5 Accuracy of 81.17\%.
Conversely, the worst model is ResNet18+Transformer, experiencing a drop of approximately 7\% compared to ResNet18+LSTM and about 6\% compared to I3D. Despite this, the Top 5 accuracies of all the models are comparable, reaching an accuracy of approximately 80\%. In light of the inherently challenging nature of the task, all three models exhibit significant overall performance.
The ResNet18+LSTM most frequently occurring errors highlight the challenging nature of the task,
including misclassification of AU2 (Outer Brow Raiser) as AU1 (Inner Brow Raiser) and confusing AU15 (Lip Corner Depressor) with AU23 (Lip Compressor). Another noteworthy mistake involves the model predicting AU26 instead of AU25, overlooking the distinction between Lip Opening and Jaw Dropping.
These discrepancies arise due to the inherent similarity between these Action Units.
We also provide a qualitative analysis of the coefficients $\alpha_{AU}$, generated by ResNet18+LSTM. In Fig. \ref{fig:qualitative_samples} we show the RGB and event frame with the corresponding 3D face shape obtained by warping a neutral identity-free face model with the regressed $\alpha_{AU}$. We color-code the 3D mesh by highlighting the distance from the neutral reference face, hence showing the most active face parts.
The ability of our approach to infer shape faces frame-by-frame thus provides a better characterization of the observed faces, as well as classifying the Action Units.
Finally, we investigate the contribution of the cross-modal loss $\mathcal{L}_{\alpha}$ by training new models without it (Tab.~\ref{tab:ablation}). The additional supervision offered by the regression task over the coefficients $\alpha_{AU}$ shows consistent improvement for the AU classification.
The positive impact of regressing the $\alpha_{AU}$ coefficients suggests that incorporating information from 3D face reconstruction helps the model better discern subtle nuances in facial movements associated with different AUs.

    


\newcommand{\meshwidth}{.15\linewidth}
\begin{figure}[t]
\centering
    \includegraphics[width=\meshwidth]{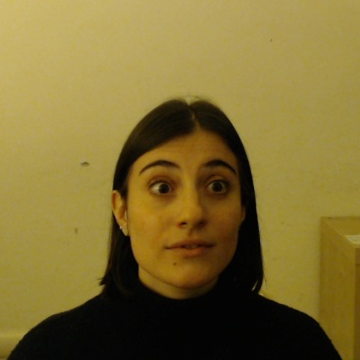}
    \includegraphics[width=\meshwidth]{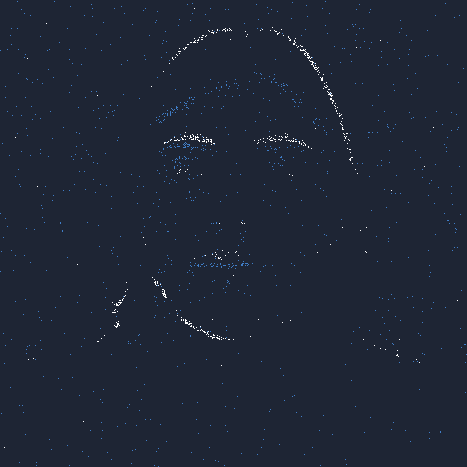}
    \includegraphics[width=\meshwidth]{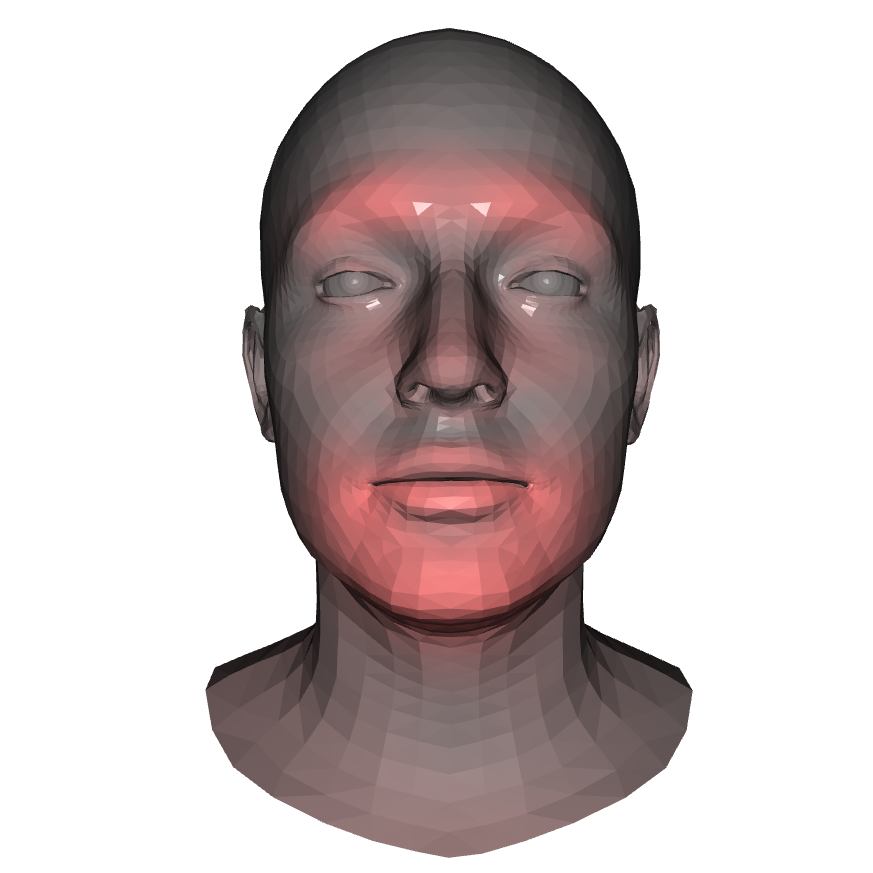}
    \includegraphics[width=\meshwidth]{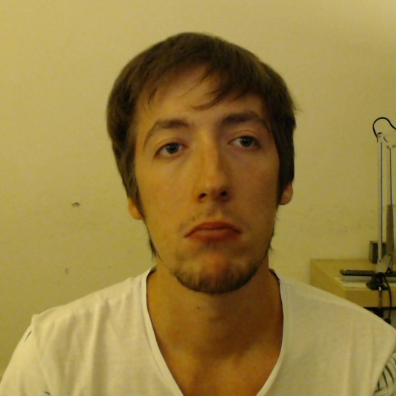}
    \includegraphics[width=\meshwidth]{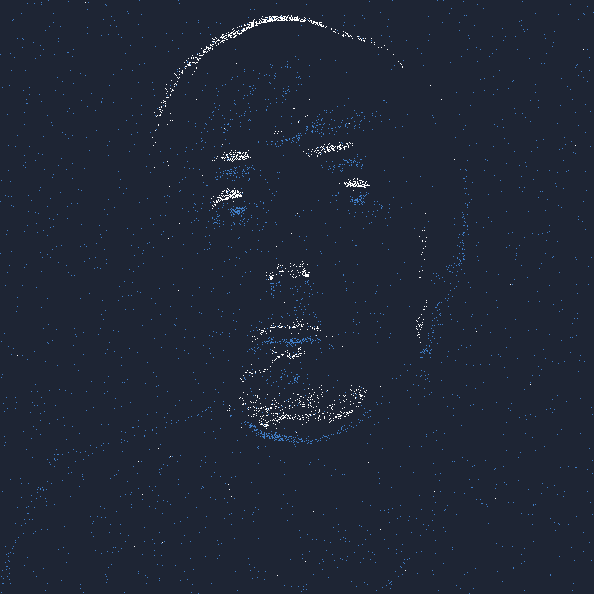}
    \includegraphics[width=\meshwidth]{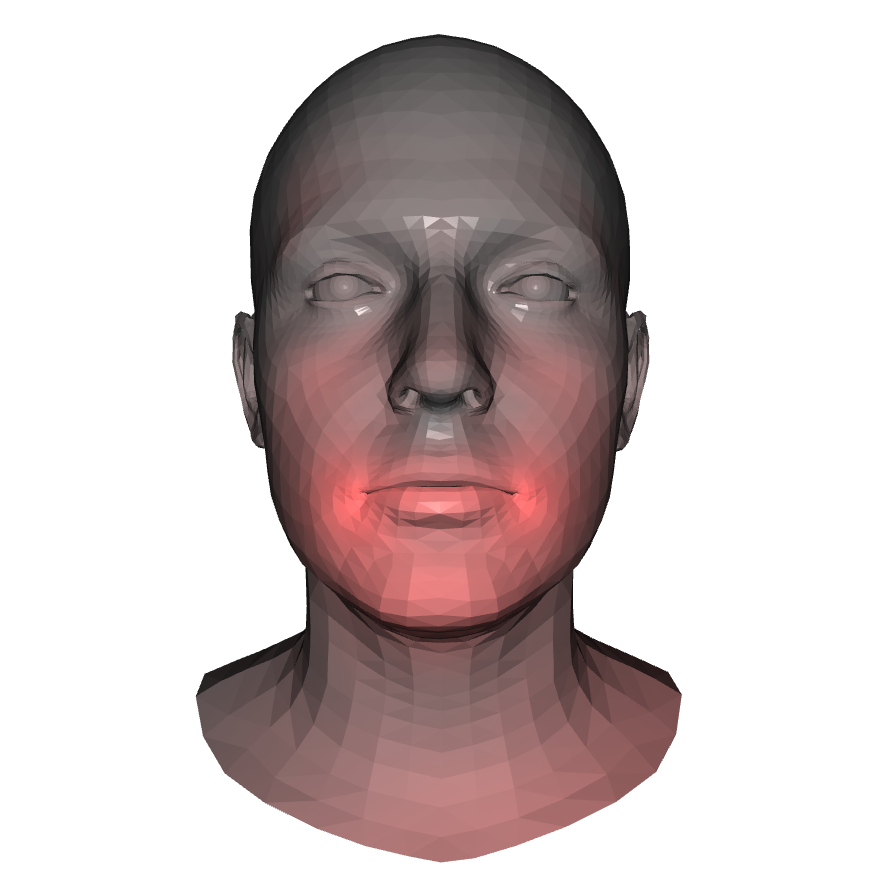}

    \includegraphics[width=\meshwidth]{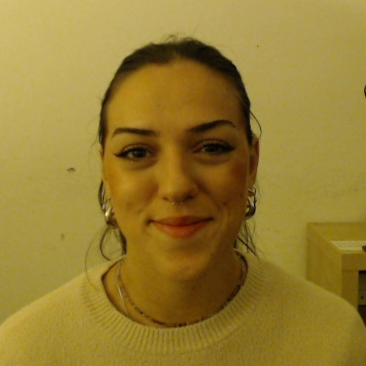}
    \includegraphics[width=\meshwidth]{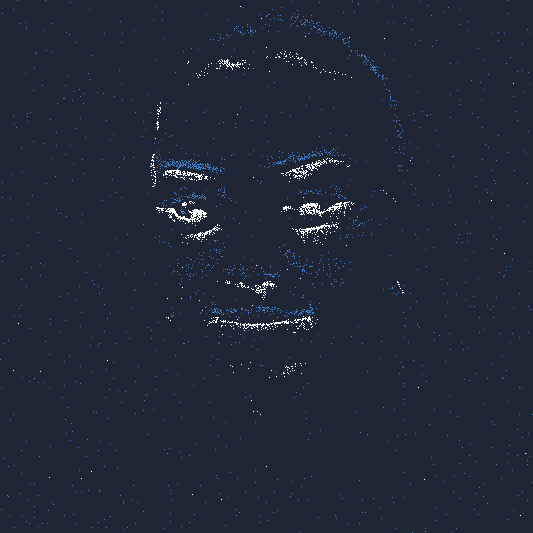}
    \includegraphics[width=\meshwidth]{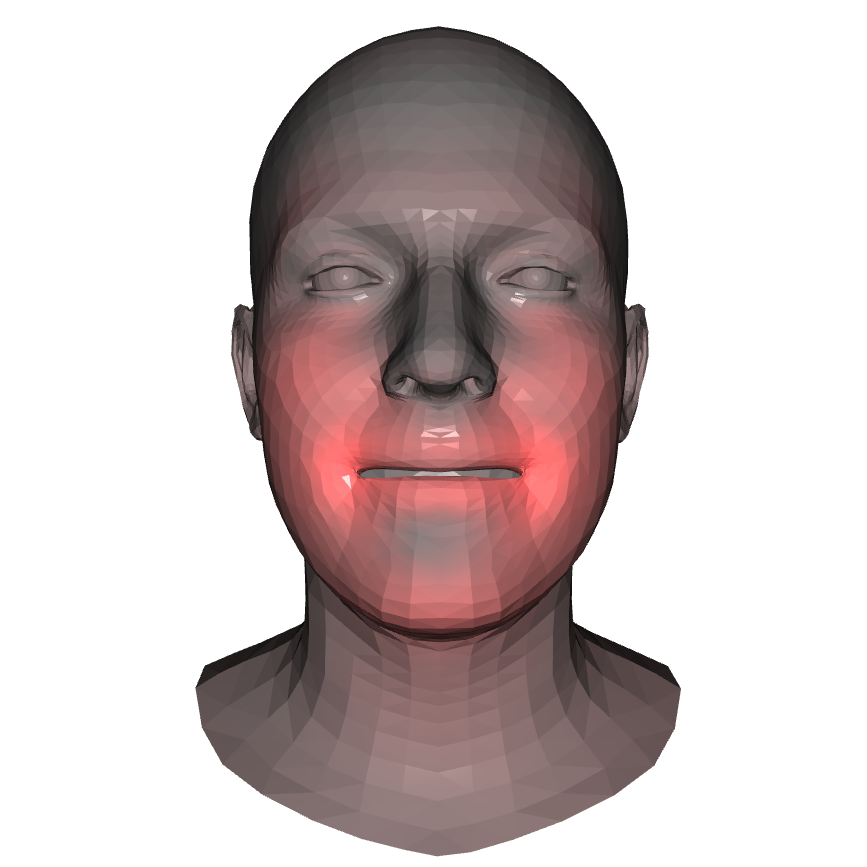}
    \includegraphics[width=\meshwidth]{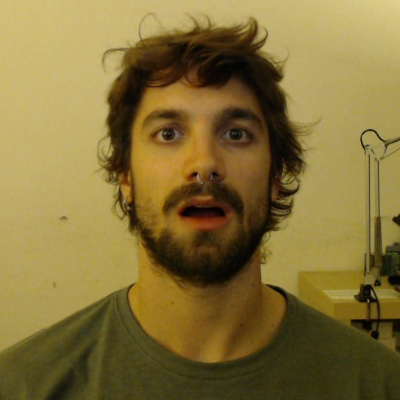}
    \includegraphics[width=\meshwidth]{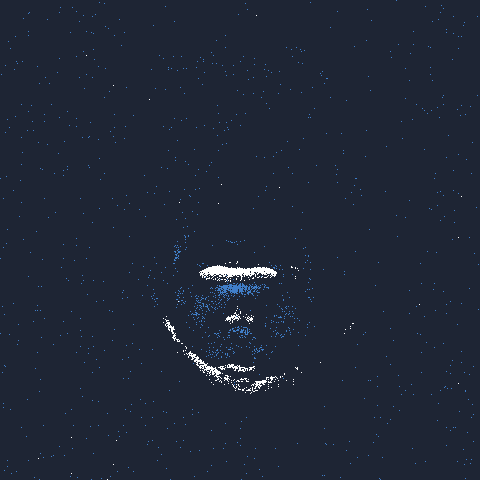}
    \includegraphics[width=\meshwidth]{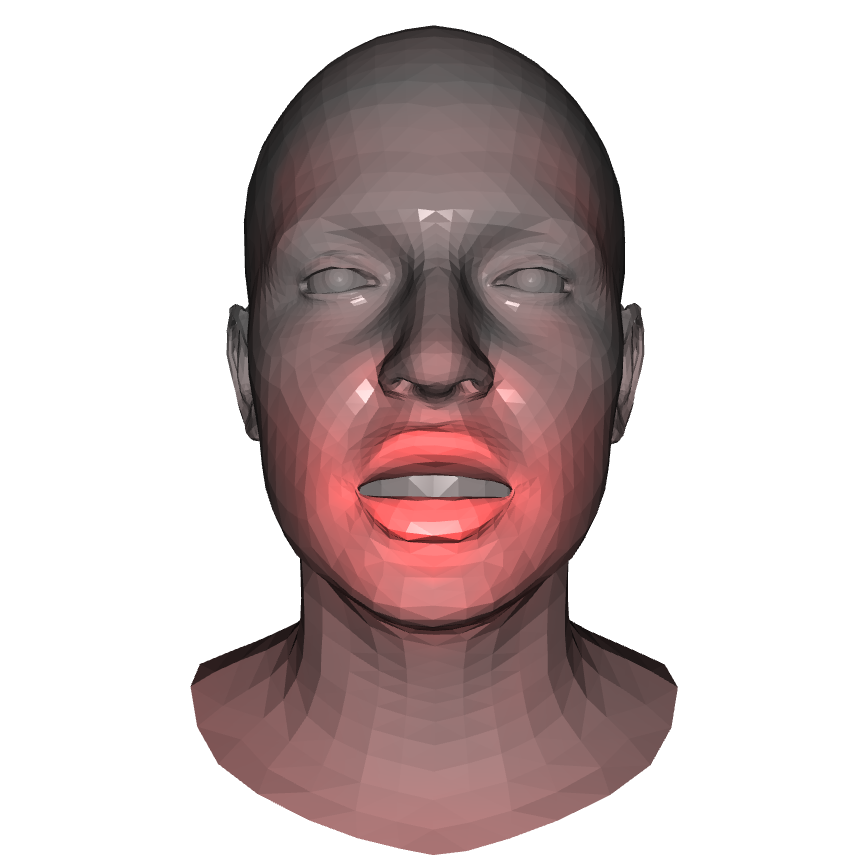}

    \caption{Samples of Action Units being performed and estimated 3D face shape. \textit{Left}: RGB frame; \textit{Center}: corresponding event frame; \textit{Right:} Reconstructed 3D mesh with the most active parts colored in \textcolor{red}{red} as distance from a neutral reference model.}
    \label{fig:qualitative_samples}
\end{figure}




\begin{table}[t]
\caption{AU classification from Event images with and without the regression loss $\mathcal{L}_{\alpha}$.}
	\centering
	\resizebox{0.8\textwidth}{!}{
		\begin{tabular}{@{}lcccc@{}}
			\toprule
			Model~~~ &  ~~~$\mathcal{L}_{\alpha}$~~~ & ~~~Accuracy~~~ & ~~~top 3 Accuracy~~~ & ~~~top 5 Accuracy~~~\\
			\midrule
   			ResNet18+LSTM & \cmark & \textbf{50.21} & 70.71 & \textbf{81.17}\\
			ResNet18+Transf. & \cmark & 43.35 & 65.59 & 80.05\\
			I3D & \cmark & 49.58 & \textbf{71.08} & 79.41\\
			\midrule
			ResNet18+LSTM & \xmark & 46.23 & 62.91 & 68.58\\
			ResNet18+Transf. & \xmark & 31.74 & 40.63 & 52.08\\
			I3D & \xmark & 47.08 & 69.58 & 80.66\\

    \bottomrule
		\end{tabular}
	}
	\label{tab:ablation}
\end{table}



\section{Conclusions and Future Works}
We have presented the \dataset{} dataset, the first event-based Action Unit classification dataset, which is paired with temporally synchronized RGB footage. To perform Action Unit classification effectively, we leveraged a cross-modal supervision by extracting pose-invariant face shape coefficients from RGB frames using a 3D Morphable Model. In our experiments, we show that regressing such coefficients frame-per-frame, while training a video-level classifier, largely improves the overall classification accuracy. As a byproduct the model can also offer a better description of the face by producing a 3D reconstruction online, as the stream is processed.
Future work directions should include the analysis of different encoding strategies for event streams, as they can heavily influence the data volume, its representation, and consequently the appropriate architecture.
Similarly, a modeling of events with finer accumulation times could offer benefits in capturing less perceivable facial movements, at the cost of a higher computational burden.

\section{Acknowledgments}
This work was partially supported by the European Commission under European Horizon 2020 Programme, grant number 951911—AI4Media. This work was partially supported by the Piano
per lo Sviluppo della Ricerca (PSR 2023) of the University of Siena - project FEATHER: Forecasting and Estimation of Actions and Trajectories for Human-robot intERactions.

\bibliographystyle{splncs04}
\bibliography{main}

\end{document}